\documentclass[runningheads]{llncs}
\usepackage{graphicx}
\usepackage{xcolor}
\usepackage{subcaption}
\usepackage{algorithm}
\usepackage{algpseudocode}
\usepackage{multirow}
\usepackage{url}
\begin{document}

\title{OCR Language Models with Custom Vocabularies}

\author{Peter Garst \inst{1}  \and
Reeve Ingle\inst{1}
\and
Yasuhisa Fujii\inst{1}
}
\authorrunning{P. Garst et al.}
%
\institute{Google, Mountain View, CA 94303, USA
\email{\{pgarst,reeveingle,yasuhisaf\}@google.com}
}

\newcommand{\todo}[1]{\textcolor{red}{TODO: #1}}
\newcommand{\todoa}[2]{\textcolor{red}{TODO(#1): #2}}

%

\maketitle              
\begin{abstract}
Language models are useful adjuncts to optical models for producing accurate optical character recognition (OCR) results. One factor which limits the power of language models in this context is the existence of many specialized domains with language statistics very different from those implied by a general language model - think of checks, medical prescriptions, and many other specialized document classes.
This paper introduces an algorithm for efficiently generating and attaching a domain specific word based language model at run time to a general language model in an OCR system. In order to best use this model the paper also introduces a modified CTC beam search decoder which effectively allows hypotheses to remain in contention based on possible future completion of vocabulary words.
The result is a substantial reduction in word error rate in recognizing material from specialized domains.
\end{abstract}

\section{Introduction}
Optical character recognition (OCR) is a fundamental tool enabling many applications, such as visual search, document digitization, understanding and translating scene text, and support for the visually impaired~\cite{nagy2000twenty,zhu2016scene,bissacco2013photoocr,thakare2018document,shen2012towards,neat2019scene}.

OCR systems range from the very general, supporting arbitrary input in many of the world's writing systems~\cite{smith2007overview,walker2018web,diaz2021rethinking}, to the very specific, for example for bank checks~\cite{jackel1995banking,chin1995microprocessor} or license plates~\cite{du2013licenseplate,agbemenu2018automatic}. The aim of this paper is to describe a system which allows general OCR systems to be quickly and easily configured for specialized tasks at run time, in many cases providing the benefits of a custom system engineered for a specific application at very small cost. The lever for making this change is the language model. 

It has been clear for decades that language models can improve OCR results by estimating the prior probability of OCR outputs~\cite{bokser1992omnidocument,smith2011limits}. Originally this was accomplished by postprocessing the output of OCR systems, in effect applying a spelling checker of some sort to the output. More recent systems often integrate language models into their decoders~\cite{YasuhisaFujii:2015:ICDAR,diaz2021rethinking,sabir2017implicit}.

Independent of their application in OCR systems, there has been a long history of creating language models with some specificity. Adaptation of speech recognition systems and handwriting recognition systems to individual users has been studied for a long time, and in many cases modifications to the associated language models has been part of that \cite{jelinek1991dynamic,chen2015recurrent}. Many of these systems create an adapted language model as a sum or interpolation of two models of the same type.

Standalone language models, used for tasks like question answering or summarization, also benefit from specialization or fine tuning \cite{caseiro2006specialized,dodge2020fine}. There is, for example, a language model trained to work well on radiology reports~\cite{yan2022radbert}.

The combination of task specific language models with successful general purpose OCR engines leads to many specialized applications, such as recognition of receipts, invoices, tax forms, medical prescriptions or notes, and many others. Some of the specialized language models used in other applications require extensive training, but in order to provide fast and simple run time configuration the models discussed here require only a vocabulary list and some frequency information for the words in the vocabulary.

The goal of this paper is to define language models and an OCR decoder architecture which efficiently solves the specialization problem for many applications. These are our contributions:
\begin{itemize}
\item We define simple language models for words and regular expressions which may contain domain specific vocabularies
\item We provide tools to generate these models from domain text, and also allow flexible user configuration
\item These models may be quickly added to existing general purpose language models at run time
\item We modify the CTC decoder to support these models with a limited kind of lookahead
\end{itemize}

\section{Custom Vocabulary Models}

\subsection{Baseline System}
Test images for this work may be either line images or full page images. In the full page image case, the full recognition system includes some preliminary material which finds text lines in the image and feeds those to a line recognizer. If the test set contains line images, then the line recognizer may consume them directly. Only the line recognizer varies between the baseline and experimental systems, so we will focus on that and treat the line segmentation code as a constant part of the environment.

The baseline line recognizer is a general purpose OCR system using a CTC \cite{graves:icml2006} beam search decoder \cite{YasuhisaFujii:2015:ICDAR}. The input line image is divided into frames (possibly with an overlapping sliding window), and at each frame the beam search maintains a list of hypotheses, each of which is an assignment of a character label to each preceding frame. In general multiple frames are mapped to one output character, with a special blank label to indicate a transition from a character to an identical character.

Each hypothesis has a score. At each frame, the decoder accepts the optical model score for each possible label in that frame. The decoder generates a new hypothesis for each preceding hypothesis and each possible label for the new frame. The score for the new hypothesis combines the score for the preceding hypothesis; the optical score for the proposed new label; the cost of a character unigram prior; transition costs for new characters, blank labels, and repeated characters; and, in the first frame for a new character, the cost of a character language model. The parameter values are optimized to minimize the character error rate on a development set with a black-box optimization \cite{46180}.

The decoder maintains a list of the best scoring hypotheses, keeping only the best $N$, and also pruning those too far away from the best. $N$ is called the beam width, and is typically 30 for the baseline recognizer.

The subject of this paper is the last component, the language model score. The baseline system includes a character based language model which estimates the probability of each possible next character in the search, given the left context of the characters already present in the hypothesis. At the frame in which the CTC search transitions to a new character, a weighted negative log probability of the new character is added to the score.
In the baseline system the same pretrained language model is applied to all input.

\subsection{Custom Vocabularies}
We wish to specify a set of words which are likely to appear in input images, and boost the score of any hypothesis in the beam search which contains one. There are a number of properties of the vocabulary to specify:

\begin{itemize}
    \item The algorithm supports both literal words and regular expressions.
    \item The literal words may be case sensitive or insensitive.
    \item Vocabulary entries may optionally be anchored to the start or end of word.
    \item Each vocabulary entry has a weight.
\end{itemize}

There are two reasons for drawing a distinction between literal words and regular expressions. The first is just implementation efficiency. The second is that the scoring algorithms, which we will discuss below, work better for fixed length vocabulary items. In the current system that is just the literal words, but this should be applied to other fixed length regular expressions as well.

As we will see the scoring algorithm includes a number of hyperparameters as well.
A vocabulary with these items specified is the essential information a user must supply, along with the input data, to benefit from this algorithm. Vocabulary sizes in tests so far have ranged from a handful of items to a few tens of thousands of items.

The tools used in the tests below can generate the vocabulary from sample text, so in general this should not be a burdensome requirement. The user is free to specify some or all of the vocabulary if there are words of particular importance in an application.

Ideally one would retune the CTC parameters and language model weights after adding a custom vocabulary, but that would not be consistent with the goal of adding new vocabularies at runtime with low latency, and in practice good results appear not to require this.

\subsection{Designing appropriate vocabularies}
In some applications the vocabulary may be clear. In processing prescriptions, the medication names are the words the user is most concerned to recognize correctly. In other applications the user may have a body of data from a specific domain, but the appropriate vocabulary is not clear.

The essential element of designing a vocabulary is choosing appropriate weights for the words. If the user already has a specific vocabulary this is all that is required. Otherwise, we may process a body of text from the domain, finding the common words and calculating their weights, and use some cutoff on the weight values to choose the vocabulary.
We use three factors in choosing the weight for a word.
\begin{enumerate}
    \item The length of the word. The scoring formula adds a value proportional to the length of the word, to give a per-character change in the score, but this does not fully capture the effect of word length. Short words tend to have many more possible confusions in the text than long words, so if short and long words have the same per-character score delta there will be more short false positives than long ones. Thus the weight for longer words should be higher. In the PubMed dataset below ``palmitoylation" is a common word which is not easily confusable with others - it benefits from a high weight.
    \item The empirical word distribution. A body of text produces an empirical probability distribution for the words. Frequent words should get higher weights.
    \item The language model distribution. We have a base language model which can estimate the probability distribution for the next character, given the left context. For each word in a body of text this leads to a language model score for the word.
\end{enumerate}

We have experimented with a number of functions of these factors and settled on a simple form:
\begin{equation}
    (c_0 + c_1 \cdot \mathit{length} + c_2 \cdot (\mathit{frequency}/\mathit{lm\_score})),
\end{equation}
where the $c_n$ are parameters chosen to minimize the OCR error rate. That is, the weight is higher for long words, and for words which are frequent in the domain text but do not score well in the baseline language model.

We have done a black-box optimization \cite{46180} to choose these parameters in a number of data sets and chosen values which work well in a variety of cases, although there is some difference in the optimal values for different kinds of tasks. Ideally one would do a fresh training for each data set, but that may not be feasible with fast run time configuration. In the future weight formulas which account for the coverage by the vocabulary of the target text and the confusability of the vocabulary items may provide more precise choice of the parameters.

\subsection{Language model state}
The language model represents regular expressions and literal words separately as finite state machines. Classes available in the OpenFST toolkit \cite{allauzen2007openfst} represent the machines.
The regular expressions in the vocabulary are separately compiled, with final state weights representing the weight of the expression in the vocabulary. These are then combined into a single state machine and optimized. Each state has two scores: one is the weight of the regular expression with that as a valid final state, if any; and the other is the best weight for which the state represents a prefix.

The literal words are handled similarly, with the vocabulary compiled into a trie, or prefix tree. Each node in the trie has two numbers attached: one is the weight of the word which ends at that node, or 0 if there is none; and the other is the weight of the best word for which the node is a prefix.
These two state machines, plus the baseline character model and a number of hyperparameters, comprise the custom vocabulary model.

Each hypothesis in the beam search includes a language model state, which depends just on the textual transcription of that hypothesis, not the way it is divided into frames. The essential properties of the state are that it can generate a score, and that appending a new character leads us to a new state.
The state includes independent components for the states of the base character model, the regular expressions, and the literal words.

Given a hypothesis which contains a sequence of characters ``abcd'' in its transcript, we must consider that a vocabulary word may start with the character a, or with the character b, and so on. Thus the decoder state for the literal words, for example, will be a vector of trie states from the trie representing the literal vocabulary. If we use Trie(``abcd") to represent the state of the trie we get by traversing the string ``abcd" from the start state, then the literal word portion of the decoder state will be [Trie(``abcd"), Trie(``bcd"), Trie(``cd"), Trie(``d")]. As we decode many of these strings will be invalid - that is, not a valid prefix for any of the vocabulary words - so the vector will in practice be pretty small.
Thus the full decoder language model state contains the state of the character model, whatever that may be; a vector of states from the literal vocabulary trie; and a vector of states from the regular expression state machine:
\begin{equation}
    (C, V_L, V_R)
\end{equation}
where $C$ is the character model state, $V_L$ is the vector of valid trie states, and $V_R$ is the vector of valid regular expression states.

If the model configuration anchors the vocabulary words to the word start position there will be many fewer valid states active at any one time. For the literal words we could also use the Aho-Corasick algorithm \cite{aho1975efficient,lee2007generalized} to generate a single state machine valid for any starting position in the string.

\subsection{Scoring the language model}

\begin{figure}[ht!]
\centering
    \includegraphics[scale=0.8]{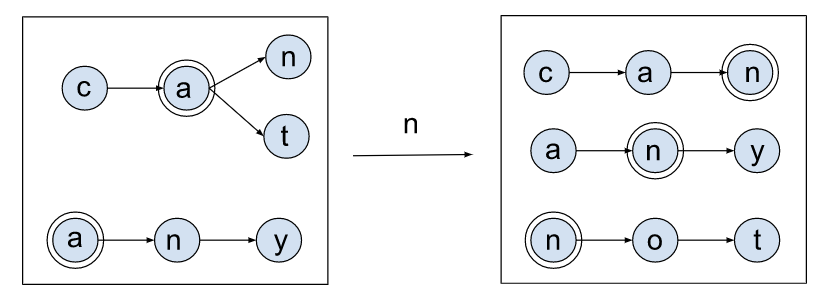}
\caption{State transition for the literal word part of the language model state. In this example the vocabulary contains can, cat, any and not. The first box is the literal word state after seeing ``ca", and the second box after we accept ``n". In the second state ``can" contributes to the base score, and all the other words in contention contribute to the best score.}
\label{fig:transition}
\end{figure}

Suppose we have a hypothesis in the beam search and we wish to transition to a new character $c$. We have the language model state for the hypothesis and the new character, and we must produce the new language model state and the score, containing both the base score for the material we have seen and the best score representing possible future word completions.

The underlying character model produces a score which is part of both base and best scores, and transitions its part of the state $C$ to a new character model state however it chooses.

For the word part of the state, each element of $V_L$ transitions to a new trie state with the addition of the character c. In some cases the transition will be invalid and the state will drop out; in the other cases the trie generates both a base and a best score. If it is consistent with the model configuration, we may also add a new state to the vector by transitioning from the start state of the trie with the character $c$. All these actions together generate the word part of the next state.

The base score for this word transition is the optimal value among all the base values for valid trie transitions. It represents the most valuable word in the vocabulary completed by the new character $c$. Similarly, the best score is the optimal value among all the best scores for the valid transitions, reflecting the most valuable partially completed vocabulary word. These word base and best scores are part of the base and best scores for the whole model. Figure \ref{fig:transition} shows an example.

The actions for the regular expression state list is similar. Adding this part in, at the end we have a new language model state, containing a new character model state and new word and regular expression state vectors; and we have base and best scores for all the parts together.

\subsection{Dual criterion beam search}

\begin{figure}[tb]
\centering
  \begin{subfigure}{0.4\textwidth}
    \includegraphics[width=\linewidth]{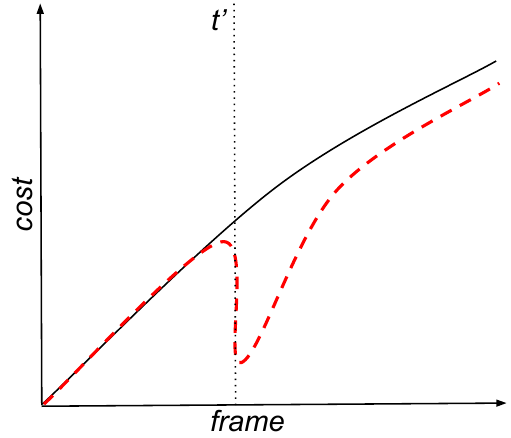}
    \caption{A boosted hypothesis is the best hypothesis at the end of search.}
    \label{fig:need_dual_beam_criterion_better}
  \end{subfigure}
  \hfill
  \begin{subfigure}{0.4\textwidth}
    \includegraphics[width=\linewidth]{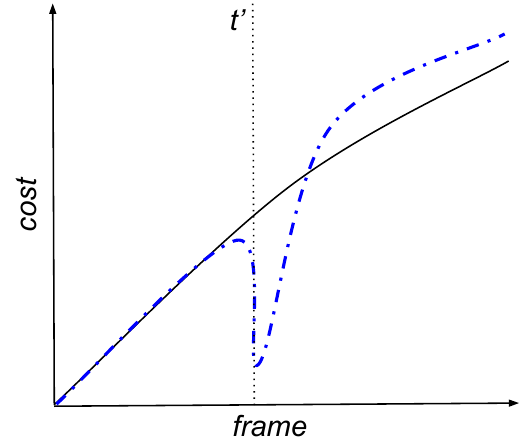}
    \caption{A boosted hypothesis is not the best hypothesis at the end of search.}
    \label{fig:need_dual_beam_criterion_worse}
  \end{subfigure}
\caption{Conceptual diagram to show the need of dual criterion beam search. The black solid line is the best hypothesis without the cost bonus. The red and blue dashed lines are hypotheses with a cost boost at frame $t'$. If we use a single criterion, the decoding can successfully find the best hypothesis for (a) but could fail for (b) because the hypotheses expanded for the boosted hypothesis can easily dominate the beam. The proposed algorithm keeps the top-$N$ and $M$ hypotheses for each case to deal with the problem.}
\label{fig:need_dual_beam_criterion}
\end{figure}

We have seen that there are two scores we attach to a hypothesis during the beam search, representing what we have seen and what we hope to see. Because beam search capacity is a limited resource, neither alone is a reliable guide to which hypotheses we should keep. If we rely only on the current scores we may eliminate a vocabulary word because of some poorly formed characters in the middle, reducing the recall on the in vocabulary words in the image. If we rely only on the hopeful scores we may force out some hypotheses with mediocre current scores, and then the hopeful hypotheses may drop out anyway if the data goes in a different direction.
Fig~\ref{fig:need_dual_beam_criterion} visualizes this problem. Thus we keep hypotheses in the beam search if the base score is good, or if the best score is good.

The dual criterion beam search is is presented in algorithm~\ref{alg:beam}.

\begin{algorithm}
\caption{Dual criterion beam search for one frame}\label{alg:beam}
\textbf{Input:} hypotheses from the previous frame, and candidate characters for this frame
\begin{enumerate}
    \item Generate a new set of hypotheses using the scoring algorithm outlined above, based on the hypotheses from the previous frame and the possible characters for the new frame. Each hypothesis will have a base score and a best score.
\item Pick out the top up to $N$ hypotheses, using the base score, subject to a constraint on the width of the beam. These will remain in the beam for the following frame. In the base system these are all the hypotheses kept by the algorithm.
\item Pick out up to $M$ additional hypotheses, based on the best score, subject to the constraint that the best score is no worse than the worst base score hypothesis in the original set of $N$. The union of these two sets is the set of hypotheses presented to the next frame.

\end{enumerate}
\end{algorithm}

Most of the benefit of the custom vocabulary comes from the language model, but in some cases the dual criterion beam search provides an additional improvement.

\subsection{Performance considerations}
This algorithm is useful in the context of a running OCR service, for which users wish to specify at run time a custom vocabulary which applies to some group of input images. As such, there are two latency figures of concern.

The additional computation required to maintain the additional beam hypotheses and to score the finite state machines associated with the custom vocabulary is negligible for common vocabulary sizes compared to the effort required to generate the optical model scores. If at some point greater efficiency becomes important the Aho-Corasick algorithm could be used to simplify and streamline the processing.

The more important latency value is for initialization - given a configuration file containing the vocabulary and associated parameters, and a running baseline OCR service, how long does it take until the service is ready to use the model? Constructing the appropriate state machines from a textual representation is straightforward, and times in the range 2 - 10 ms. on common desktop hardware are typical. This initialization time would be amortized over as many images as are used with the custom model.

\section{Experimental Results}
We explore these algorithms with a number of data sets, with different characteristics and different levels of information available. None of these data sets is perfect. The PubMed set has the most complete and accurate ground truth information, and is large enough to use better quality statistical tests in validating the algorithm.
The other data sets don't have enough information for real statistical rigor, but give at least a qualitative sense of how the algorithm performs in other situations.

\subsection{PubMed research papers}
\subsubsection{The PubMed data set}

\begin{figure}[htb!]
\centering
\includegraphics[scale=1.7]{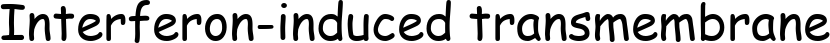}
\includegraphics[scale=1.7]{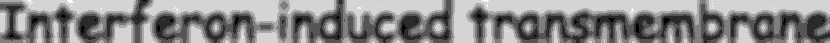}
\includegraphics[scale=1.7]{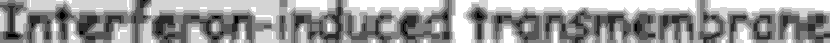}
\caption{Sample images from the PubMed dataset}
\label{fig:sample}
\end{figure}

This data set is synthetic, based on a set of related biomedical research papers from PubMed, https://pubmed.ncbi.nlm.nih.gov \cite{white2020pubmed}. IFITM3 (Interferon induced transmembrane protein 3, https://www.ncbi.nlm.nih.gov/gene/10410) \cite{sayers2021database} is implicated in the immune response to influenza, Sars-Cov-2, and other viruses, and is an active current research topic. 
We took 19 research papers from PubMed related to this topic and used the Pango (https://pango.gnome.org/) \cite{Taylor2004PangoAO} typesetter program to generate text line images with random fonts and styles from their text. We then used two different levels of random degradation on the initial images to generate images more challenging for OCR. Figure \ref{fig:sample} shows typical images generated in this manner.

This synthetic data set is not an exact model for real images one might see as OCR input, but it provides an excellent experimental platform for judging the relative efficacy of different algorithms. The original images are too easy for this test, with word error rates under 0.5\% for the baseline OCR system, but the other two versions were used for much of the development and tuning of the system.
At each degradation level the data set contains 40898 lines and 183619 words.

\subsubsection{PubMed results}

\begin{table}[hbt!]
\centering
\begin{tabular}{|c|c|c|c|c|c|c|c|c|c|c|}
\hline
\multirow{3}{*}{Vocab} &
\multirow{3}{*}{coverage} &
\multicolumn{3}{|c|}{All text} &
\multicolumn{3}{|c|}{In vocabulary} &
\multicolumn{3}{|c|}{Out of vocabulary} \\
\cline{3-11}
& & \multicolumn{2}{|c|}{WER} & \multirow{2}{*}{change} & \multicolumn{2}{|c|}{WER} & \multirow{2}{*}{change} & \multicolumn{2}{|c|}{WER} & \multirow{2}{*}{change} \\ \cline{3-4}\cline{6-7}\cline{9-10}
& & Base & Custom & & Base & Custom & & Base & Custom & \\ \hline
200 & 0.223 & \multirow{4}{*}{7.10} & 6.34 & -10.7\% & 6.00 & 1.96 & -67.4\% & 7.42 & 7.60 & 2.4\% \\
\cline{1-2}\cline{4-11}
400 & 0.287 & & 6.28 & -11.6\% & 5.14 & 1.66 & -67.7\% & 7.89 & 8.14 & 3.2\% \\
\cline{1-2}\cline{4-11}
800 & 0.365 & & 6.19 & -12.9\% & 4.51 & 1.45 & -67.8\% & 8.59 & 8.91 & 3.8\% \\
\cline{1-2}\cline{4-11}
1200 & 0.424 & & 6.12 & -13.9\% & 4.29 & 1.39 & -67.6\% & 9.17 & 9.60 & 4.7\% \\
\hline
\end{tabular}
\caption{Heavily degraded PubMed recognition results as a function of vocabulary size}
\label{tab:pubmed1}
\end{table}

\begin{table}[hbt!]
\centering
\begin{tabular}{|c|c|c|}
\hline
Vocab size &
All text ratio &
In vocabulary ratio \\
\hline
200 & 3.62 & 14.45 \\
\hline
400 & 3.74 & 14.48 \\
\hline
800 & 4.10 & 14.88 \\
\hline
1200 & 4.44 & 14.83 \\
\hline
\end{tabular}
\caption{Heavily degraded PubMed win ratios}
\label{tab:pubmed1a}
\end{table}

\begin{table}[hbt!]
\centering
\begin{tabular}{|c|c|c|c|c|c|c|c|c|c|c|}
\hline
\multirow{3}{*}{Vocab} &
\multirow{3}{*}{Coverage} &
\multicolumn{3}{|c|}{All text} &
\multicolumn{3}{|c|}{In vocabulary} &
\multicolumn{3}{|c|}{Out of vocabulary} \\
\cline{3-11}
& & \multicolumn{2}{|c|}{WER} & \multirow{2}{*}{change} & \multicolumn{2}{|c|}{WER} & \multirow{2}{*}{change} & \multicolumn{2}{|c|}{WER} & \multirow{2}{*}{change} \\ \cline{3-4}\cline{6-7}\cline{9-10}
& & Base & Custom & & Base & Custom & & Base & Custom & \\ \hline
200 & 0.223 & \multirow{4}{*}{1.65} & 1.47 & -11.0\% & 1.32 & 0.16 & -88.2\% & 1.74 & 1.85 & 6.3\% \\
\cline{1-2}\cline{4-11}
400 & 0.287 & & 1.46 & -11.7\% & 1.08 & 0.14 & -86.9\% & 1.88 & 1.99 & 5.9\% \\
\cline{1-2}\cline{4-11}
800 & 0.365 & & 1.46 & -11.8\% & 0.86 & 0.12 & -86.5\% & 2.10 & 2.23 & 6.2\% \\
\cline{1-2}\cline{4-11}
1200 & 0.424 & & 1.46 & -11.6\% & 0.75 & 0.11 & -84.7\% & 2.31 & 2.45 & 6.1\% \\
\hline
\end{tabular}
\caption{Lightly degraded PubMed recognition results as a function of vocabulary size}
\label{tab:pubmed2}
\end{table}

\begin{table}[hbt!]
\centering
\begin{tabular}{|c|c|c|}
\hline
Vocab size &
All text ratio &
In vocabulary ratio \\
\hline
200 & 4.50 & 22.64  \\
\hline
400 & 4.72 & 17.53  \\
\hline
800 & 4.52 & 17.10  \\
\hline
1200 & 4.11 & 13.62  \\
\hline
\end{tabular}
\caption{Lightly degraded PubMed win ratios}
\label{tab:pubmed2a}
\end{table}

\begin{figure}[hbt!]
\centering
\includegraphics[scale=1.7]{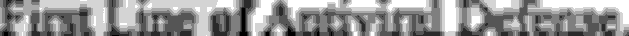}
\caption{A win: corrected ``First Line of Antivirel Defenso."}
\label{fig:wins}
\end{figure}

\begin{figure}[hbt!]
\centering
\includegraphics[scale=1.7]{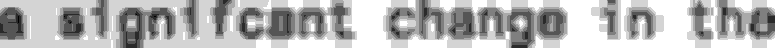}
\includegraphics[scale=1.7]{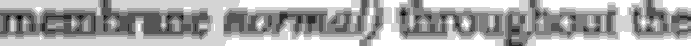}
\caption{Losses: changed correct ``signifcant" to ``significant" and ``normal" to ``formal"}
\label{fig:loss}
\end{figure}

Table~\ref{tab:pubmed1} shows figures of merit for these models as a function of vocabulary size. These results used the more heavily degraded version of the data set. These figures were generated using a jackknife protocol, with one paper at a time left out, and the vocabulary automatically generated. At each vocabulary size we see the vocabulary coverage, the word error rates and the relative change in error rate. The coverage is the portion of all the words in the document which are in vocabulary. The figure includes these values for the entire documents, and just for the in vocabulary words or out of vocabulary words. We see that there is some price in terms of lower accuracy for out of vocabulary words, but a much stronger positive effect for the in vocabulary and whole document sets.

Note that the in vocabulary baseline accuracy changes with different vocabulary sizes. This is because the set of in vocabulary words changes from line to line. Each word will get exactly the same baseline results on each line, but the set of included words changes.

Table~\ref{tab:pubmed1a} shows the win ratio at each vocabulary size for the heavily degraded data, for all the text and for the in vocabulary text. The win ratio is the ratio of the number of errors corrected by the model to the number of words which were correct in the base model but changed to an incorrect value in the custom vocabulary model.

The model configuration anchored the vocabulary items to word start, and a word is considered in vocabulary if it benefits from the algorithm. For example, if ``with" is in the vocabulary, then ``within" is considered an in vocabulary word in the data set.

Tables~\ref{tab:pubmed2} and \ref{tab:pubmed2a} show the same results with the more lightly degraded version of the data set, with baseline word error rate 1.65. The baseline in vocabulary error rate varies with the vocabulary size, but is generally lower than the overall error rate.

Figures \ref{fig:wins} and \ref{fig:loss} show some wins and losses on this data set.

\subsection{Handwritten prescriptions}

\begin{figure}[hbt!]
\centering
\includegraphics[scale=0.8]{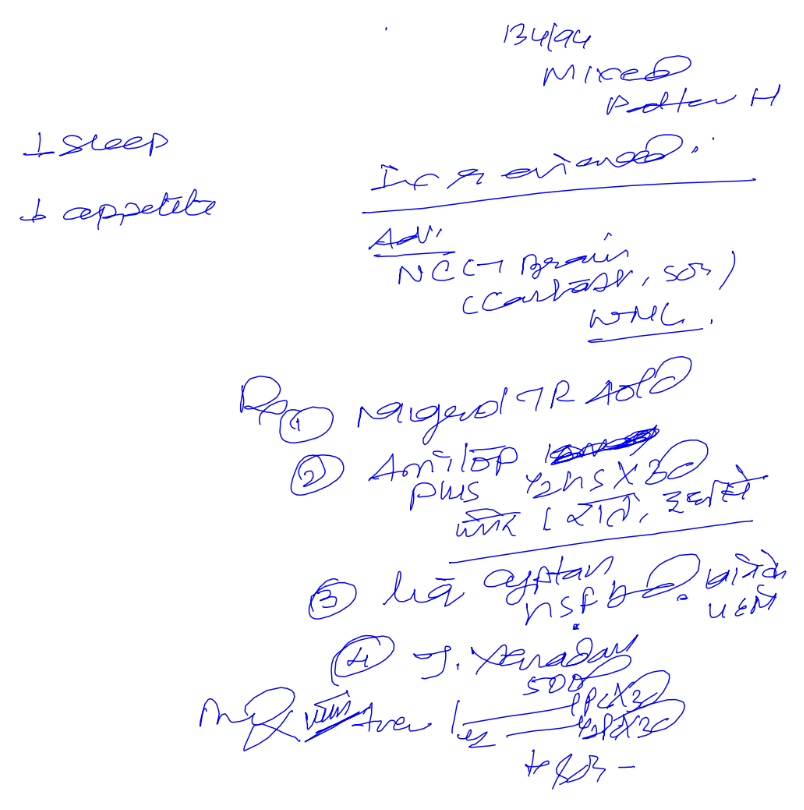}
\caption{This prescription contains amitop, migrol, xenadom and cyptan}
\label{fig:written}
\end{figure}

\subsubsection{The prescription data set}
This data set contains full page images of handwritten prescriptions.  A sample is shown in figure \ref{fig:written}. The ground truth for these images contains only the list of medications mentioned in each prescription, and not the other text elements or the locations of the medication words. This limits the available options for analyzing OCR performance on these images, but the Jaccard Index can be used as an important measure of algorithm quality. 

The Jaccard Index will be a perfect 1 if the OCR system finds all the medications in a prescription, without introducing any false positives. Given an image, if $GT$ is the set of ground truth vocabulary words in the image, and $R$ is the set of recognized vocabulary words in the image, then the index is

\begin{equation}
J(R,GT) = \frac{| R \cap GT|}{| R \cup GT |}
\end{equation}

The Jaccard Index has important limitations, but it is an appropriate metric for this application. It provides no information on location or order, but none is required in the output. This data set also has the property that each vocabulary word appears at most once in an image, which is helpful in interpreting the metric.

The vocabulary contains 41844 words, consisting of names of medications, with 6111 of them actually used in at least one image. The data set contains 9647 images, with 47712 in vocabulary words altogether.
\subsubsection{Prescriptions results}

\begin{table}[hbt!]
\centering
\begin{tabular}{|r@{ }|c|}
\hline
\multicolumn{1}{|c|}{Vocabulary size} & Jaccard Index \\
\hline
0 & 0.193  \\
\hline
500 & 0.201   \\
\hline
1000 & 0.209  \\
\hline
5000 & 0.271   \\
\hline
10000 & 0.257   \\
\hline
41844 & 0.178  \\
\hline
\end{tabular}
\caption{Handwritten prescription recognition results as a function of vocabulary size}
\label{tab:prescrip}
\end{table}

For this data set we have for each image a list of the medicine names in the image, and we use the Jaccard Index as the figure of merit for the model. This is a challenging data set, containing images of messy handwritten prescriptions. As noted, this data set contains full page images rather than line images, so the error rates may include segmentation errors as well as recognition errors, but the segmentation material is the same in all experiments so we will ignore it.

This data set allows us to further explore the results of vocabulary size. The initial size is 41844 words. This includes some synonyms and abbreviations, but is larger than any list of commonly prescribed medications - the mobile version of the Physician's Desk Reference, for example, has about 2500 medications. The images actually use 6111 of the medication names.

Table~\ref{tab:prescrip} shows the Jaccard Index achieved by this model as a function of vocabulary size. For this experiment, to test vocabulary size $V$ we split the input images into 5 parts. We used a word list generated by the first four parts to create a model for evaluating the fifth part. The list contains vocabulary words actually used in the first 4 parts, filled out to size $V$ by adding random words from the master list. If more than $V$ words are actually used in the first 4 parts, we select them by the weight as discussed above. The values in the table are pooled from evaluating all 5 data slices in this manner.

In this data set we see that including too many words is counterproductive, and better results are obtained by focusing on the common words. We might achieve better results if we had truly accurate frequency information for the whole vocabulary.

\subsection{Medicine names}
This data set contains medicine names cropped from the handwritten prescription data set. This is a useful set for comparing the original single beam search with the dual beam search. We see in Table~\ref{tab:medname} that the language model is doing the heavy lifting, reducing the case insensitive word error rate from 47.45 to 7.64. In this application the dual beam search provides a further 16\% drop in word error rate. 

\begin{figure}[hbt!]
\centering
\includegraphics[scale=1.5]{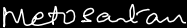}
\caption{Fixed by the dual beam search: Meto Pantan $\rightarrow$ Metosartan}
\label{fig:fixed}
\end{figure}

Figure~\ref{fig:fixed} shows an example of a word which was fixed by the dual beam search. This makes sense - the split between the right and wrong versions happens well before the end of the word, and so in the single beam search there are many opportunities to prune the correct result before the end. In other applications, like the pay stub data set discussed below, the dual beam search is not helpful.

\begin{table}[hbt!]
\centering
\begin{tabular}{|c|c|c|c|}
\hline
Base WER &
Single beam search &
Dual beam search &
Change due to dual beam search \\
\hline
47.45 & 7.64 & 6.43 & -16\% \\
\hline
\end{tabular}
\caption{Effect of the dual beam search on case insensitive word error rate}
\label{tab:medname}
\end{table}

\subsection{Pay stubs}
\begin{figure}[hbt!]
\centering
\includegraphics[scale=0.3]{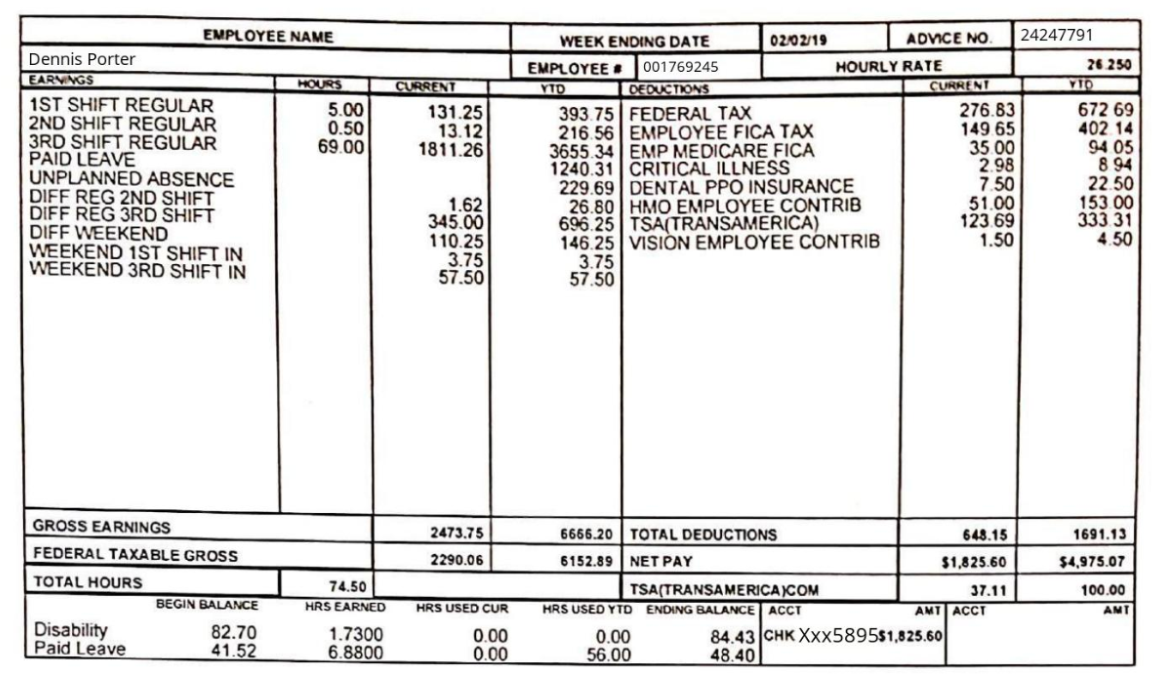}
\caption{Sample pay stub}
\label{fig:paystub}
\end{figure}

This data set contains images of pay stubs; figure~\ref{fig:paystub} shows a typical example.

For this data set we used a custom vocabulary containing only a few regular expressions representing different ways of expressing monetary amounts and dates. One could also reasonably add relevant words, like taxable, federal, deductions and so on, and perhaps further improve the results.

\begin{figure}[hbt!]
\centering
\includegraphics[scale=0.3]{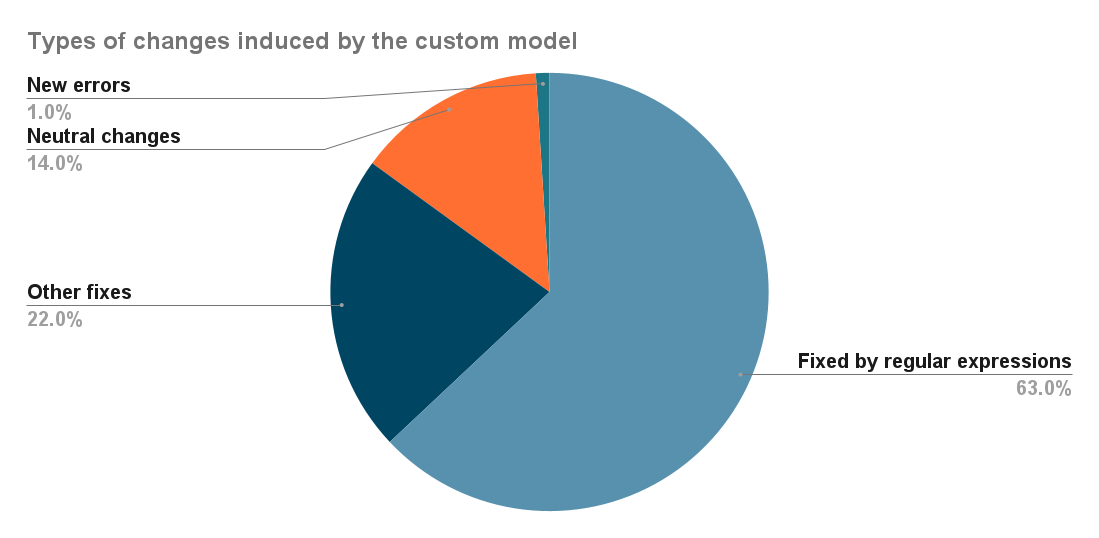}
\caption{Changes induced by the custom pay stub model}
\label{fig:paychart}
\end{figure}

Because we do not have ground truth for this data set we ran the base model and the custom vocabulary model and examined a random sample of all the words where the results differed between the two models; the results are shown in figure~\ref{fig:paychart}. We do not have error rates for either set of recognition results, but since there were 63 instances of words being fixed by the regular expressions against 1 new error, it is clear the results were broadly positive. Many of the fixes were humble but nonetheless useful, for example correcting confusions between commas and decimal points.

The other fixes category contains things that could have been fixed with a more tightly configured base model, and so do not really redound to the credit of the custom vocabulary. For example, ``PTO" might be interpreted as three Latin characters or three Cyrillic characters by the base model, but this custom vocabulary model was constrained to Latin script output.

\subsection{Retail price tags}
\subsubsection{The price tags data}
The next data set contains real images of price tags attached to items in a store. The vocabulary for this data is somewhat limited, containing a lot of brand names and a number of common phrases (``for a limited time only"). There are also many elements of the images well described by regular expressions, like monetary amounts or quantities (``32 oz.").
The data set contains 101 images with 385 annotated lines.

\subsubsection{Price tags results}
The price tag task is different from the previous ones in several respects.
\begin{itemize}
    \item The vocabulary covers essentially the whole document.
    \item Regular expressions are an important part of the vocabulary.
    \item Because of the small data set size, the training text for vocabulary statistics is the same as the test set text.
\end{itemize}

Because of the last item, the results are optimistic, but are suggestive of results one could obtain in the field.

The vocabulary included a few hand selected regular expressions; future work may allow us to generate them automatically from sample text, as we do the literal word lists. They are:

\begin{itemize}
    \item \textbackslash \$ \textbackslash d+
    \item \textbackslash d+\textbackslash .\textbackslash d
    \item \textbackslash d+ ?(CT $\vert$ LB $\vert$ OZ $\vert$ EA $\vert$ ML $\vert$ MG)
\end{itemize}

Table~\ref{tab:tags} shows the results. We see that almost all the changes are corrections. This table also shows the effects of the dual criterion beam search. The two beam search columns use the same custom vocabulary language model, but the first uses the baseline single criterion beam search, and the second the dual criterion beam search.

\begin{table}[hbt!]
\centering
\begin{tabular}{|c|c|c|c|}
\hline
\multicolumn{3}{|c|}{WER} & \multirow{2}{*}{win ratio} \\ \cline{1-3}
Baseline & Single Beam Search & Dual Beam Search & \\ \hline
15.16 & 10.93 & 10.45 & 24.5 \\ \hline

\end{tabular}
\caption{Recognition results on price tags}
\label{tab:tags}
\end{table}

\section{Conclusions}
We observed that in a variety of data sets from different domains, language models incorporating custom domain specific vocabularies may be leveraged to substantially improve the accuracy of optical character recognition models. The language models support both literal words and regular expressions, and have a number of configuration options which enhance flexibility.

We introduced algorithms which permit these custom models to be automatically derived from a body of text in the target domain. Users may also partially or completely design their own models in special cases, for example if there is a list of key words which the user is particularly concerned to recognize correctly.

We also introduced a modified CTC decoder to support these models which in effect provides in-vocabulary lookahead in order to use information about partially completed as well as complete words to improve accuracy.

The models discussed here introduce no significant overhead to the recognition process, and they may be added to an OCR service at run time with low latency.

Future work will aim to further improve the algorithms for combining the models. It would also be useful to better address low information situations, in effect adapting to specialized input streams rather than designing a model based on prior knowledge about a domain. 

\bibliographystyle{splncs04}
\bibliography{references}

\begin{thebibliography}{10}
\providecommand{\url}[1]{\texttt{#1}}
\providecommand{\urlprefix}{URL }
\providecommand{\doi}[1]{https://doi.org/#1}

\bibitem{agbemenu2018automatic}
Agbemenu, A.S., Yankey, J., Addo, E.O.: An automatic number plate recognition
  system using opencv and tesseract ocr engine. International Journal of
  Computer Applications  \textbf{180}(43), ~1--5 (2018)

\bibitem{aho1975efficient}
Aho, A.V., Corasick, M.J.: Efficient string matching: an aid to bibliographic
  search. Communications of the ACM  \textbf{18}(6),  333--340 (1975)

\bibitem{allauzen2007openfst}
Allauzen, C., Riley, M., Schalkwyk, J., Skut, W., Mohri, M.: Openfst: A general
  and efficient weighted finite-state transducer library: (extended abstract of
  an invited talk). In: Implementation and Application of Automata: 12th
  International Conference, CIAA 2007, Praque, Czech Republic, July 16-18,
  2007, Revised Selected Papers 12. pp. 11--23. Springer (2007)

\bibitem{bissacco2013photoocr}
Bissacco, A., Cummins, M., Netzer, Y., Neven, H.: Photoocr: Reading text in
  uncontrolled conditions. In: Proceedings of the ieee international conference
  on computer vision. pp. 785--792 (2013)

\bibitem{bokser1992omnidocument}
Bokser, M.: Omnidocument technologies. Proceedings of the IEEE  \textbf{80}(7),
   1066--1078 (1992)

\bibitem{caseiro2006specialized}
Caseiro, D., Trancoso, I.: A specialized on-the-fly algorithm for lexicon and
  language model composition. IEEE Transactions on Audio, Speech, and Language
  Processing  \textbf{14}(4),  1281--1291 (2006)

\bibitem{chen2015recurrent}
Chen, X., Tan, T., Liu, X., Lanchantin, P., Wan, M., Gales, M.J., Woodland,
  P.C.: Recurrent neural network language model adaptation for multi-genre
  broadcast speech recognition. In: Sixteenth Annual Conference of the
  International Speech Communication Association (2015)

\bibitem{chin1995microprocessor}
Chin, F., Wu, F.: A microprocessor-based optical character recognition check
  reader. In: Proceedings of 3rd International Conference on Document Analysis
  and Recognition. vol.~2, pp. 982--985. IEEE (1995)

\bibitem{diaz2021rethinking}
Diaz, D.H., Qin, S., Ingle, R., Fujii, Y., Bissacco, A.: Rethinking text line
  recognition models. arXiv preprint arXiv:2104.07787  (2021)

\bibitem{dodge2020fine}
Dodge, J., Ilharco, G., Schwartz, R., Farhadi, A., Hajishirzi, H., Smith, N.:
  Fine-tuning pretrained language models: Weight initializations, data orders,
  and early stopping. arXiv preprint arXiv:2002.06305  (2020)

\bibitem{du2013licenseplate}
Du, S., Ibrahim, M., Shehata, M., Badawy, W.: Automatic license plate
  recognition (alpr): A state-of-the-art review. IEEE Transactions on Circuits
  and Systems for Video Technology  \textbf{23}(2),  311--325 (2013).
  \doi{10.1109/TCSVT.2012.2203741}

\bibitem{YasuhisaFujii:2015:ICDAR}
Fujii, Y., Genzel, D., Popat, A.C., Teunen, R.: Label transition and selection
  pruning and automatic decoding parameter optimization for time-synchronous
  {V}iterbi decoding. In: Proceedings of the 13th International Conference on
  Document Analysis and Recognition. pp. 756--760. IEEE (Aug 2015)

\bibitem{46180}
Golovin, D., Solnik, B., Moitra, S., Kochanski, G., Karro, J.E., Sculley, D.
  (eds.): Google Vizier: A Service for Black-Box Optimization (2017),
  \url{http://www.kdd.org/kdd2017/papers/view/google-vizier-a-service-for-black-box-optimization}

\bibitem{graves:icml2006}
Graves, A., Fern\'{a}ndez, S., Gomez, F., Schmidhuber, J.: Connectionist
  temporal classification: Labelling unsegmented sequence data with recurrent
  neural networks. In: ICML (2006)

\bibitem{jackel1995banking}
Jackel, L.D., Sharman, D., Stenard, C.E., Strom, B.I., Zuckert, D.: Optical
  character recognition for self-service banking. AT\&T Technical Journal
  \textbf{74}(4),  16--24 (1995). \doi{10.1002/j.1538-7305.1995.tb00189.x}

\bibitem{jelinek1991dynamic}
Jelinek, F., Merialdo, B., Roukos, S., Strauss, M.: A dynamic language model
  for speech recognition. In: Speech and Natural Language: Proceedings of a
  Workshop Held at Pacific Grove, California, February 19-22, 1991 (1991)

\bibitem{lee2007generalized}
Lee, T.H.: Generalized aho-corasick algorithm for signature based anti-virus
  applications. In: 16th International conference on computer communications
  and networks. pp. 792--797. IEEE (2007)

\bibitem{nagy2000twenty}
Nagy, G.: Twenty years of document image analysis in pami. IEEE Transactions on
  Pattern Analysis and Machine Intelligence  \textbf{22}(1),  38--62 (2000)

\bibitem{neat2019scene}
Neat, L., Peng, R., Qin, S., Manduchi, R.: Scene text access: A comparison of
  mobile ocr modalities for blind users. In: Proceedings of the 24th
  International Conference on Intelligent User Interfaces. pp. 197--207 (2019)

\bibitem{sabir2017implicit}
Sabir, E., Rawls, S., Natarajan, P.: Implicit language model in lstm for ocr.
  In: 2017 14th IAPR international conference on document analysis and
  recognition (ICDAR). vol.~7, pp. 27--31. IEEE (2017)

\bibitem{sayers2021database}
Sayers, E.W., Beck, J., Bolton, E.E., Bourexis, D., Brister, J.R., Canese, K.,
  Comeau, D.C., Funk, K., Kim, S., Klimke, W., et~al.: Database resources of
  the national center for biotechnology information. Nucleic acids research
  \textbf{49}(D1), ~D10 (2021)

\bibitem{shen2012towards}
Shen, H., Coughlan, J.M.: Towards a real-time system for finding and reading
  signs for visually impaired users. ICCHP (2)  \textbf{7383},  41--47 (2012)

\bibitem{smith2007overview}
Smith, R.: An overview of the tesseract ocr engine. In: Ninth international
  conference on document analysis and recognition (ICDAR 2007). vol.~2, pp.
  629--633. IEEE (2007)

\bibitem{smith2011limits}
Smith, R.: Limits on the application of frequency-based language models to ocr.
  In: 2011 International Conference on Document Analysis and Recognition. pp.
  538--542. IEEE (2011)

\bibitem{Taylor2004PangoAO}
Taylor, O.: Pango, an open-source unicode text layout engine (2004)

\bibitem{thakare2018document}
Thakare, S., Kamble, A., Thengne, V., Kamble, U.: Document segmentation and
  language translation using tesseract-ocr. In: 2018 IEEE 13th International
  Conference on Industrial and Information Systems (ICIIS). pp. 148--151. IEEE
  (2018)

\bibitem{walker2018web}
Walker, J., Fujii, Y., Popat, A.C.: A web-based ocr service for documents. In:
  Proceedings of the 13th IAPR international workshop on document analysis
  systems (DAS), Vienna, Austria. vol.~1 (2018)

\bibitem{white2020pubmed}
White, J.: Pubmed 2.0. Medical reference services quarterly  \textbf{39}(4),
  382--387 (2020)

\bibitem{yan2022radbert}
Yan, A., McAuley, J., Lu, X., Du, J., Chang, E.Y., Gentili, A., Hsu, C.N.:
  Radbert: Adapting transformer-based language models to radiology. Radiology:
  Artificial Intelligence  \textbf{4}(4),  e210258 (2022)

\bibitem{zhu2016scene}
Zhu, Y., Yao, C., Bai, X.: Scene text detection and recognition: Recent
  advances and future trends. Frontiers of Computer Science  \textbf{10},
  19--36 (2016)

\end{thebibliography}

\end{document}